\documentclass[letterpaper, 10 pt, conference]{ieeeconf}  

\IEEEoverridecommandlockouts                              

\usepackage[pdftex]{graphicx}
\usepackage{graphicx}
\usepackage{graphics} 
\usepackage{amsmath} 
\usepackage{amssymb}  
\usepackage{CJK}
\usepackage{algorithm}
\usepackage{algorithmicx}
\usepackage{algpseudocode}
\usepackage{cite}
\usepackage{booktabs} 

\title{\LARGE \bf
Learning Gating ConvNet for Two-Stream based Methods in Action Recognition
}

\author{Jiagang Zhu$^{1,2}$, Wei Zou$^{1}$, Zheng Zhu$^{1,2}$
\thanks{$^{1}$Institute of Automation, Chinese Academy of Sciences, Beijing,
People{\textquoteright}s Republic of China, {\tt\small \{zhujiagang2015, wei.zou, zhuzheng2014\}@ia.ac.cn}}%
\thanks{$^{2}$University of Chinese Academy of Sciences, Beijing, People{\textquoteright}s
Republic of China}
\thanks{This work is supported in part by the
National Natural Science Foundation of China under Grant
No. 61403378, and in part by the National
High Technology Research and Development Program of
China under Grant No.2015AA042307.}
}

\begin{document}

\newcommand{\tabincell}[2]{\begin{tabular}{@{}#1@{}}#2\end{tabular}}
\maketitle
\thispagestyle{empty}
\pagestyle{empty}

\begin{abstract}
For the two-stream style methods in action recognition, fusing the two streams\textquoteright~predictions is always by the weighted averaging scheme. This fusion method with fixed weights lacks of pertinence to different action videos and always needs trial and error on the validation set. In order to enhance the adaptability of two-stream ConvNets and improve its performance, an end-to-end trainable gated fusion method, namely gating ConvNet, for the two-stream ConvNets is proposed in this paper based on the MoE (Mixture of Experts) theory. The gating ConvNet takes the combination of feature maps from the same layer of the spatial and the temporal nets as input and adopts ReLU (Rectified Linear Unit) as the gating output activation function. To reduce the over-fitting of gating ConvNet caused by the redundancy of parameters, a new multi-task learning method is designed, which jointly learns the gating fusion weights for the two streams and learns the gating ConvNet for action classification. With our gated fusion method and multi-task learning approach, a high accuracy of 94.5\% is achieved on the dataset UCF101.
\end{abstract}

\section{Introduction}
Human action recognition is important for applications of human-robot interaction, behavior analysis and surveillance. Early works \cite{Laptev05,IDT13,MOFAP16} utilized hand-crafted spatial-temporal local descriptors and powerful encoding methods for video representations and classification. Inspired by the successes of deep learning for image classification \cite{ImageNet12}, lots of works have explored deep convolutional neural networks (CNN) \cite{twostream14,TSN16} for video classification and achieved higher performance than hand-crafted methods recently.

\begin{figure}[t]
\centering
\includegraphics[width=1\linewidth]{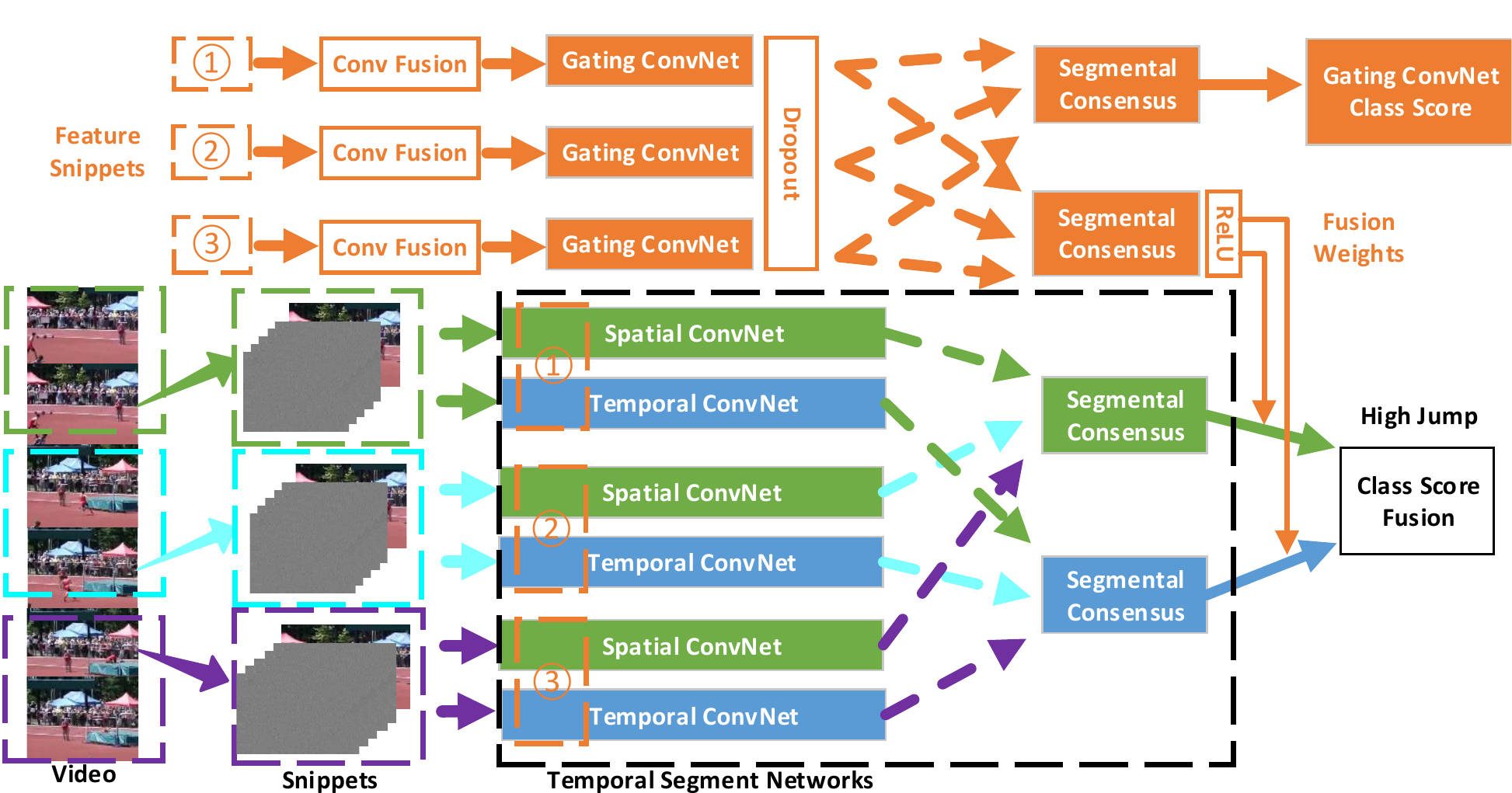}
\caption{Gated TSN: Newly added gating ConvNet are in orange color. Its inputs are called feature snippets, where number ${\textcircled{1}}, {\textcircled{2}}, {\textcircled{3}}$ denote the different segment level feature maps from the two streams. Each of feature snippets is the combination of feature maps from the same layer of the spatial and the temporal nets. There are two independent fully connected layers for the gating ConvNet, one for learning the gating fusion weights for the two streams, the other for the action classification of the gating ConvNet.}
\label{Figure1}
\end{figure}

This paper mainly focuses on improving the performance of the two-stream ConvNets in action recognition. The two-stream ConvNets \cite{twostream14} contain the spatial net and the temporal net, which take RGB frames and consecutive optical flow stacks as inputs respectively. The predictions of the two streams are always fused by evenly averaging or weighted averaging \cite{twostream14,TSN16}. Though Wang {\sl et al}. \cite{TSN16} has given more credits to the temporal net due to its higher accuracy than the spatial one, this fixed weight fusion method cannot make the best use of the capacity of the spatial and the temporal nets. Because each of them fires on different aspects of videos: the spatial net focuses on the appearance and scene contents of videos, while the temporal one on the motion. Also, different video frames of the same and the different classes contain different amount of spatial and temporal cues. Fusing the predictions of the spatial and the temporal nets with fixed weight may not capture the contents of videos well and always needs trial and error on the validation set. Some fusion methods have also been proposed for the two-stream ConvNets, but they are not designed for fusing the predictions between the two streams. The SCI (Sparsity Concentration Index) fusion \cite{SunJYS15} gives a weighted score scheme according to the sparsity degrees of the crop-level prediction. However, their spatial stream and temporal stream are combined by the concatenation of feature maps of the last convolutional layers. The SCI fusion in their work is used to fuse the predictions of different crops from a single spatio-temporal stream, while our work is for fusing the predictions of the two streams. The conv fusion \cite{convfusion16} also fuses the two streams in feature level to get a spatio-temporal stream and a temporal stream. The spatio-temporal stream is constructed by concatenating two groups of feature maps of the spatial and the temporal nets and then fusing this new group of feature maps by 3D convolution and 3D pooling. But they fuse the predictions of the spatio-temporal and temporal stream by evenly averaging.

In this paper, a gating network \cite{Jacobs91} is developed to obtain the adaptive fusion weights for the spatial and the temporal nets in prediction level. We gain insights from the Mixture of Experts (MoE) \cite{Jacobs91}, which is a popular method for function approximation. In the standard architecture of MoE, there are usually a gating network and more than two expert networks. Each expert network is gated via a Softmax function of the gating network. Shazeer {\sl et al}. \cite{ShazeerMMDLHD17} proposed a Sparsely-Gated Mixture-of-Experts layer to maintain the model capacity without proportionally increasing computation cost. Their MoE consists of up to thousands of feed-forward sub-networks and they use a gating network to determine a sparse combination of experts for each sample. Localized gating \cite{Ramamurti98} improves the generalization capacity of the MoE by growing and shrinking the number of experts according to the model performance on the validation set. However, there are some differences between the above MoE architectures and our work. Firstly, different from these MoE methods with more than two experts, our work aims to best utilize the spatial and the temporal nets of the two-stream based action recognition methods. Using three or more experts is not in the scope of this paper. Secondly, the inputs for different experts in previous MoE methods are usually from the same source or a small subset of the same source. While the gating network proposed in this paper takes the combination of convolutional feature maps of the spatial and the temporal nets as inputs. It is termed as {\bf gating ConvNet} for it is mainly composed of several convolutional layers. Our fusion method based on the gating ConvNet is termed as {\bf gated fusion}. Thirdly, instead of choosing Softmax as the final gating output activation function of the gating ConvNet, ReLU is employed to output two non-negative fusion weights for the two streams in prediction level. The gating ConvNet with ReLU as the output activation function shows faster convergence in network training and competitive accuracy in final testing than that with Softmax. Finally, a new multi-task learning method is proposed, which jointly learns the adaptive fusion weights for the two streams and the gating ConvNet for action classification. Different from the weighted averaging fusion with fixed weights, the gated fusion is a sample specific fusion method because the gating ConvNet makes a reasonable assignment in two fusion weights for the spatial and the temporal nets adaptively according to the properties of different video inputs.

The main contributions of this paper can be summarized as follows: 1) An end-to-end trainable gated fusion method is proposed for the two-stream ConvNets. 2) A new multi-task learning method is designed, which jointly learns the gating fusion weights for the two streams and the gating ConvNet for action classification. With this multi-task learning approach, the accuracy of our MoE is improved by more than 0.1\%. 3) With our gated fusion method and multi-task learning approach, a high accuracy of 94.5\% is achieved on the dataset UCF101. Our learned model are also visualized, including the output and the feature of the gating ConvNet. Some specific classification examples are also shown to illustrate the gated fusion has better perfomances than the weighted averaging scheme with fixed weight.

\section{Related Work}
Previous works related to ours mainly fall into three categories: (1) Two-stream ConvNets for action recognition, (2) Mixture of Experts, (3) Multi-task learning.

{\bf Two-stream ConvNets for action recognition}. The two-stream \cite{twostream14} ConvNets contain the spatial and the temporal nets. By combining pre-training on ImageNet and further fine-tuning on action classification datasets \cite{ucf101, hmdb51}, this method achieves comparative performance with IDT (Improved Dense Trajectory) \cite{IDT13}, a state-of-the-art hand-crafted action recognition method. The video-level training has been explored in Temporal Segment Networks (TSN) \cite{TSN16} by averaging the segment-level predictions into the final prediction during training. While the training of the original two-stream ConvNets is in frame and short clip level, TSN adopts the video-level training by sparsely sampling frames and optical flow stacks from different parts of a video. Based on the two-stream approach, a lot of feature fusion \cite{convfusion16} and encoding \cite{TLE16, ActionVLAD17} methods are proposed. In most of these works, fusion of the two streams’ predictions is always implemented by evenly averaging or weighted averaging with fixed weights. In this paper, we consider learning the gating ConvNet for the prediction level fusion of the two streams, which falls short of research in the community of action recognition.

{\bf Mixture of Experts}. Classical Mixture of Experts (MoE) \cite{Jacobs91} contains a gating network and several expert networks. The expert networks map the input $X$ to the output $Y$, while the gating network produces a probability distribution over all experts{\textquoteright}~final predictions. For deep neural networks, Softmax is usually adopted as the gating output activation function to get a convex combination of all experts{\textquoteright}~outputs \cite{Ramamurti98} or to obtain a weighted mask in segmentation \cite{AdapNet17}. In our MoE method with two experts (spatial and temporal net), the outputs of gating ConvNet perform as the confidence ratio between the spatial net and the temporal net. ReLU can perform the same role as Softmax because of its non-negativity. Besides, ReLU has shown its faster convergence than many other activation functions in deep learning \cite{ImageNet12}. In this paper ReLU is chosen as the gating output activation function. To our best knowledge, this has not been done before.
In previous works of MoE, the inputs for different experts are from the same kind of data. This is different from the two-stream ConvNets, whose inputs are RGB frames and optical flow stacks for the spatial net and the temporal net respectively. Instead, the gating ConvNet in this paper takes the combination of feature maps of the two streams as inputs.

{\bf Multi-task learning}. Multi-task learning is a useful regularizer which could reduce over-fitting and improve performance in deep learning. In the community of object detection, Faster R-CNN \cite{fasterrcnn15} employs multi-task learning both in RPN (Region Proposal Network) training and Fast R-CNN \cite{fastrcnn15} training to do object classification and bounding box regression simultaneously. While in action recognition, the two-stream ConvNets \cite{twostream14} method combines different action datasets \cite{ucf101,hmdb51} together in the training stage and back-propagates gradients through two different classification branches which share the same input feature. In this way, over-fitting is reduced by increasing the amounts of training data. In this work, a different multi-task learning approach is proposed for action recognition, namely, learning the gating fusion weights for the two streams and learning the gating ConvNet for action classification jointly. With this multi-task learning approach, the accuracy of our MoE is improved by more than 0.1\%.

\section{Approach}
In this section, the framework of the gated fusion method will be introduced. Then three aspects of learning the gating ConvNet are detailed.

\subsection{Gated Fusion for the two-stream ConvNets}
The spatial and the temporal nets of TSN are selected to act as two experts in our basic framework for their simple architectures and good performance in the two-stream ConvNets. The reader can refer to \cite{TSN16} for more details of TSN. Fig. \ref{Figure1} shows our {\bf gated TSN}. It contains TSN and gating ConvNet. The gating ConvNet takes the combination of convolutional feature maps of the two streams as inputs. Besides, it has two independent fully connected layers, one for learning the gating fusion weights for the two streams and the other for action classification. So they can work in a multi-task learning manner. Through segmental consensus \cite{TSN16} of the gating ConvNet, the video-level fusion weights and the video-level gating ConvNet predictions for action recognition are obtained. The gating ConvNet outputs the gating fusion weights as follows

\begin{equation}
\label{eq1}
\begin{split}
& {\bf G}_g = \mathcal{H}_g(\mathcal{G}_g(\mathcal{F}_g(f_1;{\bf W}_g),F_g(f_2;{\bf W}_g),...,F_g(f_K;{\bf W}_g)))
\end{split}
\end{equation}
where $f_k,~k=1,...,K$ is the feature snippets generated by the combination of feature maps of the spatial and the temporal nets. $K$ is number of segments. $\mathcal{F}_g(f_k;{\bf W}_g)$ is the function representing the gating ConvNet with parameters ${\bf W}_g$ which operates on the feature snippet $f_k$. $\mathcal{G}_g$ aggregates the frame level fusion weights to get the video-level fusion weights. Average pooling is adopted for $\mathcal{G}_g$. $\mathcal{H}_g$ is a ReLU function ensuring the non-negativity of video-level fusion weights. With the gated fusion method, adaptive weighted function (\ref{eq2}) of the two streams is obtained, where $w_1$ and $w_2$ are the fusion weights outputted by the gating ConvNet $w_1={{G}_g}_1, w_2={{G}_g}_2$. It is also worth noting that the predictions of the two streams ${\bf G}_{rgb}$ and ${\bf G}_{flow}$ are fused before Softmax normalization \cite{TSN16, UntrimmedNets17} in our gated fusion method
\begin{equation}
\label{eq2}
{\bf G}_{adap} = w_1{\bf G}_{rgb} + w_2{\bf G}_{flow}
\end{equation}
where ${\bf G}_{adap}$ is the weighted prediction.

For the classification branch of the gating ConvNet, the prediction function is defined as

\begin{equation}
\label{eq1}
\begin{split}
& {\bf G}_c = \mathcal{G}_c(\mathcal{F}_c(f_1;{\bf W}_c),\mathcal{F}_c(f_2;{\bf W}_c),...,\mathcal{F}_c(f_K;{\bf W}_c))
\end{split}
\end{equation}
where $\mathcal{F}_c(f_k;{\bf W}_c)$ is the function representing the gating ConvNet with parameters ${\bf W}_c$ which operates on the feature snippet $f_k$. Note that ${\bf W}_c$ and ${\bf W}_g$ share parameters except for the fully connected layers. $\mathcal{G}_c$ aggregates frame level predictions into the video-level predictions. Average pooling is adopted for $\mathcal{G}_c$. The final loss function for the gating ConvNet is

\begin{equation}
\label{eq3}
\begin{split}
& {\bf L} = \mathcal{L}(y,{\bf G}_{adap}) + {\lambda}\mathcal{L}(y,{\bf G}_c)  \\
& = -\sum\limits_{i=1}^{C}y_i\Bigg({G}_{adap_i} - \log\sum\limits_{j=1}^{C}\exp{{G}_{adap_j}}\Bigg)\\
& -\lambda\sum\limits_{i=1}^{C}y_i\Bigg({{G}_c}_i - \log\sum\limits_{j=1}^{C}\exp{{{G}_c}_j}\Bigg) \\
\end{split}
\end{equation}
where $C$ is the number of action classes and $y_i$ the ground truth label of class $i$. $\mathcal{L}(y,{\bf G}_{adap})$ is the loss with respect to the predictions of the two streams after gated fusion. $\mathcal{L}(y,\bf{G}_c)$ is the loss of the classification branch of the gating ConvNet. $\lambda$ is the loss weight for classification loss of the gating ConvNet. Standard cross-entropy loss is employed for these two losses respectively. In the back-propagation process, the gradients of the gating ConvNet parameters ${\bf W}$ with respect to the loss can be derived as

\begin{equation}
\label{eq4}
\begin{split}
& \frac{\partial{{\bf L}}}{\partial{{\bf W}}} = \frac{\partial{\mathcal{L}(y,{\bf G}_{adap})}}{\partial{{\bf W}_g}} + {\lambda}\frac{\partial{\mathcal{L}(y,{\bf G}_c)}}{\partial{{\bf W}_c}} = \frac{\partial{\mathcal{L}}}{\partial{{\bf G}_{adap}}}\\
&\Bigg({\bf G}_{rgb} \frac{\partial{{\bf G}_{adap}}}{\partial{{{G}_g}_1}}\frac{\partial{{{G}_g}_1}}{\partial{{\mathcal{H}_g}}}\frac{\partial{\mathcal{H}_g}}{\partial{\mathcal{G}_g}}\sum\limits_{k=1}^{K}\frac{\partial{\mathcal{G}_g}}{\partial{\mathcal{F}_g(f_k)}}\frac{\partial{\mathcal{F}_g(f_k)}}{\partial{{ W}_g}} \\
& +{\bf G}_{flow}\frac{\partial{{\bf G}_{adap}}}{\partial{{{G}_g}_2}}\frac{\partial{{{G}_g}_2}}{\partial{{\mathcal{H}_g}}}\frac{\partial{\mathcal{H}_g}}{\partial{\mathcal{G}_g}}\sum\limits_{k=1}^{K}\frac{\partial{\mathcal{G}_g}}{\partial{\mathcal{F}_g(f_k)}}\frac{\partial{\mathcal{F}_g(f_k)}}{\partial{{ W}_g}}\Bigg) \\
& + \lambda\frac{\partial{\mathcal{L}}}{\partial{{\bf G}_c}}\sum\limits_{k=1}^{K}\frac{\partial{\mathcal{G}_c}}{\partial{\mathcal{F}_c(f_k)}}\frac{\partial{\mathcal{F}_c(f_k)}}{\partial{{\bf W}_c}}
\end{split}
\end{equation}

\subsection{Implementations of learning the gating ConvNet}
\label{learning_gating}
In this subsection, three aspects of learning the gating ConvNet will be detailed.

{\bf Output activation function for the gating ConvNet}. In our MoE method with $N = 2$ experts (spatial and temporal net), the outputs of gating ConvNet perform as the confidence ratio between the two streams. To model gating outputs $g$ as a function of its inputs ${\bf x}$, different functions could be considered. ReLU
\begin{equation}
\label{eq_relu}
g({x}_i) = max(0,{x}_i), i=1,2
\end{equation}
is selected as the output activation function because of its non-negativity and fast convergence speed \cite{ImageNet12}. It can perform the same role as Softmax \cite{Jacobs91}
\begin{equation}
\label{eq_softmax}
g({x}_i) = \frac{exp({x}_i)}{\sum_{i=1}^{N}exp({x}_i)}, i=1,2
\end{equation}
ReLU could make sense as long as the two output fusion weights $g({x}_i), i=1,2$ of the gating ConvNet do not become zero together for a specific sample.

{\bf Inputs for the gated TSN}. The inputs for the spatial and the temporal nets of the gated TSN follow the original TSN, where RGB frames and optical flow stacks are randomly sampled from each of $K$ segments of a video. Each group of RGB frames and optical flow stack starts from the same point in a video. As for the gating ConvNet, RGB frames and optical flow stacks are not taken directly as inputs. This is for two reasons. One is that these two modalities are from different distributions and training the gating ConvNet may be difficult. The other is the training of gating ConvNet also needs the forward pass of the spatial and the temporal nets. If we add layers for extracting features of RGB frames and optical flow stacks for the gating ConvNet, the memory consumption would be serious. Thus, the gating ConvNet in this paper takes the feature maps of different layers of two streams with {\bf concatenation fusion} or {\bf conv fusion} \cite{convfusion16} as inputs. Concatenation fusion ${\bf y}^{cat}~=~f^{cat}({\bf x}^a,{\bf x}^b)$ stacks the two feature maps ${\bf x}^a,{\bf x}^b$ from the two streams at the same spatial locations $i,j$ across the feature channels $d$

\begin{equation}
\label{eq_concat}
y_{i,j,2d}^{cat}=x_{i,j,d}^{a}~~~~y_{i,j,2d-1}^{cat}=x_{i,j,d}^{b}
\end{equation}
where ${\bf y}^{cat}\in\mathbb{R}^{H{\times}W{\times}2D}$. Conv fusion ${\bf y}^{conv}~=~f^{conv}({\bf x}^a,{\bf x}^b)$ first stacks the two feature maps ${\bf x}^a,{\bf x}^b$ at the same spatial locations $i,j$ across the feature channels $d$ as above equation (\ref{eq_concat}) and subsequently convolves the stacked data with a bank of filters ${\bf f}^{1\times1}\in\mathbb{R}^{1\times1\times2D{\times}D}$ and biases ${\bf b}\in\mathbb{R}^{D}$

\begin{equation}
\label{eq_conv}
{\bf y}^{conv}={\bf y}^{cat}*{\bf f}^{1\times1}+{\bf b}
\end{equation}
where ${\bf y}^{conv}\in\mathbb{R}^{H{\times}W{\times}D}$. The filter ${\bf f}^{1\times1}$ is used to reduce
the dimensionality of concatenation feature maps by a factor of two and is able to model
weighted combinations of the two feature maps from two streams at the same spatial location. Layers with different depth are chosen to get best result, which are expected to extract low-level features like textures, edges and colors in the lower layers and highly semantic features like objects in the higher layers.

{\bf Multi-Task Learning for the gating ConvNet}. The gating ConvNet takes the convolutional layers of BN-Inception \cite{BN15} after the above mentioned input fusion layer as its feature extractor, followed by a dropout layer and two independent fully connected layers. For example, if the inception3c of the spatial and the temporal nets are fused as the inputs of the gating ConvNet, the inception4a, b, c, d, e and the inception5a, b of BN-Inception would be chosen sequentially as the convolutional layers of the gating ConvNet. Note that the same feature extractor of action classification is used for the gating ConvNet. However, the dimension of classification output (101 for UCF101) is much larger than the dimension of the gating fusion weights (in the case of two-stream, 2). Thus, the gating ConvNet is equipped with redundant degrees of freedom in its feature extractor. Learning this task could be cumbersome \cite{AdapNet17} and suffers from severe risk of over-fitting. To relieve over-fitting, a action classification branch is added on top of the final convolutional layer of the gating ConvNet and it could behave as a regularizer for the task of learning the gating fusion weights. At this point, the gating ConvNet has two independent fully connected layers, one for the gating fusion weights, the other for the action classification. These two fully connected layers share the same input layer (inception5b). It is expected that joint learning of the fusion weights and classification could improve the accuracy of our MoE.

\section{Experiments}
In this section, the evaluation dataset and the implementation details of our approach are firstly introduced. Then, we do ablation studies for learning the gating ConvNet. Moreover, the performance of our method is compared with the state of the art. Finally, our learned gating ConvNet models are visualized.

\subsection{Dataset and Implementation Details}
Experiments are conducted on the standard action dataset: UCF101 \cite{ucf101}. The UCF101 dataset contains 101 action classes and 13,320 video clips. Three training/testing splits are used for evaluation. All experiments are implemented with Caffe \cite{caffe14} and one NVIDIA GTX TITAN X GPU is used for training and testing. Codes will be available at {\sl \thanks{https://github.com/zhujiagang/gating-ConvNet-code}}.

\begin{algorithm}[t]
\caption{ Training Gated TSN.}
\label{alg1}
\begin{algorithmic}[1]
\Require
\State TSN fine-tuned on action dataset;
\State The gating ConvNet pretrained on ImageNet.
\Ensure
  Gated TSN;
\State Fix the parameter of TSN.
\Repeat
\State Learning gating fusion weights;
\label{code:fram:extract}
\Until{val accuracy no more increase}
\Repeat
\State Joint learning of gating fusion weights and gating ConvNet for action classification.
\label{code:fram:trainbase}
\Until{val accuracy no more increase}
\end{algorithmic}
\end{algorithm}

{\bf Network Training}. The same training strategies in \cite{TSN16} are adopted for the TSN in our gated TSN, including cross modality pre-training, partial BN, dropout and data augmentation. The number of the snippets $K$ is set to 3 for both TSN and gating ConvNet. The loss weight $\lambda$ is set to 0 when we only learn gating fusion weights, and is set to 1 when we jointly learn gating fusion weights and the gating ConvNet for classification. For the gating ConvNet, the ImageNet pre-training is used and the dropout ratio of dropout out layer is set to 0.8. The mini-batch SGD algorithm is used to learn the network parameters. Alg. \ref{alg1} shows the training procedures of the gated TSN, which mainly include three stages: 1) Firstly the two streams of the TSN are trained; 2) Then parameters of these two streams are fixed and we only fine-tune the gating ConvNet for learning the gating fusion weights; 3)When there is no more increase in accuracy, lastly we do joint learning of the gating fusion weights and gating ConvNet for action classification. Training the gating ConvNet consumes much more memory than training the spatial and temporal nets respectively, so a smaller batch size is needed (set to 4) than the first training stage (set to 32). L2 norm of gradients is clipped at 40 and momentum term is set to 0.9. For training the gating ConvNet, the learning rate is initialized as 0.001 and decreases to 0.0001 when there is no more increase in accuracy. The model is selected by early stopping. Optical flows are extracted by the TVL1 optical flow algorithm \cite{TVL1}. RGB frames and optical flows are extracted from videos in advance.

{\bf Network Testing}. For each video during testing, 25 RGB frames and optical flow stacks are sampled. Meanwhile, the crops of 4 corner and 1 center, and their horizontal flippings are obtained from sampled frames. Each pair of RGB frame and optical flow stack starts from the same point in a video. For each pair of them, the gated fusion for the spatial and temporal net is applied. All predictions of crops in a video are averaged to get a video-level result.
\subsection{Ablation Studies}
In this section, different aspects for learning the gating ConvNet described in Sec. \ref{learning_gating} are investigated by experiments, which include selection of different output activation functions, testing of different input layers and different ways of fusing these layers. Besides, different network architectures and different two-stream methods are also implemented for our gated fusion framework. Moreover, our proposed multi-task learning approach is verified. These experiments are all conducted on the split 1 of UCF101 dataset, unless otherwise specified.

{\bf ReLU or Softmax}. For the output activation function of the gating ConvNet, two activation functions are explored: Softmax and ReLU. Feature maps with concatenation fusion or conv fusion \cite{convfusion16} of inception4e from two streams are taken as the inputs of the gating ConvNet. As shown in Table \ref{table1}, the gating ConvNet with ReLU is comparable to that with Softmax in the accuracy of the gated TSN. It can been seen that ReLU could perform the same role as Softmax when used as the final gating activation function. It is also found that the gating ConvNet with ReLU as the final activation function converges faster than that with Softmax. This complies with the fact that ReLU could have faster training speed than saturating nonlinearities when used as the layer activation function \cite{ImageNet12}. In later experiments, ReLU is adopted for the gating output activation function.

\begin{table}[t]
\caption{Accuracy (\%) of the gated TSN with Softmax and ReLU as the output activation function of the gating ConvNet respectively on the UCF101 (split 1).}
\begin{center}
\label{table1}
\vbox{
\renewcommand{\baselinestretch}{1.25}
{\footnotesize\centerline{\tabcolsep=10pt\begin{tabular*}{0.4\textwidth}{ccc}
\toprule
Input for gating ConvNet & Softmax & ReLU\\
\hline
\tabincell{c}{Conv fusion \\  of inception4e}   &94.05& \bf{94.11}\\
\tabincell{c}{Concatenation fusion \\  of inception4e} &93.97&93.96\\
\bottomrule
\end{tabular*}}}}
\end{center}
\end{table}

{\bf Different input layers and ways of fusing these layers}. As we want to collect information from the spatial and the temporal nets, feature maps from different parallel convolutional layers of the two streams are combined as the inputs of the gating ConvNet. Specifically, each pair of convolutional layers of the two streams (conv1, conv2, inception3c, inception4e, inception5b) with concatenation or conv fusion is chosen for each experiment. The performance of different input layers for the gating ConvNet is summarized in Table \ref{table2}. While different fusion methods (conv, concatenation) and different input layers perform slightly different, the inception4e as the input layer of the gating ConvNet with conv fusion gets the highest accuracy. This is different from the previous work about two-stream feature fusion \cite{convfusion16}, where the fusion of the highest convolutional layers after ReLU gains the best result. Because in \cite{convfusion16}, the fusion of the input layers is for a completely independent spatio-temporal stream and the discriminative feature from the highest convolutional layer is important. While the fusion of layers for the gating ConvNet begins from the middle layer of the two streams and the outputs of the gating ConvNet end at the predictions of the two streams. In this case, different input features and fine-tuning of the layers before the highest convolutional layer are needed. In latter experiments, the feature maps of the inception4e from the two streams with conv fusion are adopted for the inputs of the gating ConvNet, unless otherwise specified.

\begin{table}[t]
\caption{Accuracy (\%) of the gated TSN when different input layers with concatenation or conv fusion are used for the inputs of the gating ConvNet on the UCF101 dataset (split 1).}
\begin{center}
\label{table2}
\vbox{
\renewcommand{\baselinestretch}{1.25}
{\footnotesize\centerline{\tabcolsep=10pt\begin{tabular*}{0.475\textwidth}{ccc}
\toprule
\tabincell{c}{Input layer for \\ gating ConvNet}  & concatenation fusion & conv fusion\\
\hline
conv1        &93.97&93.96\\
conv2        &93.91&93.54\\
inception3c  &93.93&93.89\\
inception4e  &93.96&\bf{94.11}\\
inception5b  &93.89&93.87\\
\bottomrule
\end{tabular*}}}}
\end{center}
\end{table}

{\bf Different network architectures}. To test the generality of our gated fusion for the two-stream ConvNets in action recognition, we also do experiments on different network architectures including CaffeNet \cite{caffe14} and VGG16 \cite{vgg}. Two kinds of two-stream based methods, namely the original two-stream ConvNets \cite{twostream14} and TSN \cite{TSN16} are implemented with these architectures. The number of the snippets $K$ during training is set to 3 for TSN and 1 for the original two-stream ConvNets. Different fusion methods such as weighted averaging fusion with fixed weight, fusion based on SCI (Sparsity Concentration Index) \cite{SunJYS15} and our gated fusion method are used for all these networks. Results are summarized in Table \ref{table4}. For weighted averaging fusion, the predictions of the two streams before Softmax normalization are fused and the best weight is selected with grid search on the validation set for each experiment. As the method in \cite{SunJYS15} with SCI fusion has only one stream, in our two-stream method with SCI fusion, average fusion is added in stream level after its crop level probability fusion. For the spatial and the temporal nets with CaffeNet and VGG16, it is found that these two networks suffer from severe over-fitting in UCF101 due to the limited training data. The gated fusion can not do better than the weighted averaging fusion in already over-fitting expert networks \cite{Ramamurti98}. To reduce the over-fitting of the CaffeNet and VGG16 in the two-stream ConvNets, all their fully connected layers are removed. A dropout layer with dropout ratio of 0.8 is added after the last convolutional layer and a new fully connected layer is added for action recognition. Then the training schemes of the original two-stream and TSN are followed to get the final spatial and temporal models. For the gating ConvNets of CaffeNet and VGG16, the layers before the last convolutional layers from the two streams with conv fusion are taken as inputs. As shown in Table \ref{table4}, the gated fusion always performs the best in different architectures and different two-stream methods, which shows the advantage of assigning the gating fusion weights for the two streams. The SCI fusion performs comparably well with the weighted averaging fusion. From CaffeNet, VGG16 to BN-Inception, the classification accuracy increases, indicating the network architecture is important for complex video classification. It is also noted that TSN always outperforms the two-stream method both in CaffeNet and VGG16, demonstrating the merits of the video-level training of TSN.

{\bf Multi-task learning for the gating ConvNet}. Further, a different fully connected layer for action classification is added on top of the last convolutional layer of the gating ConvNet. The network is fine-tuned on the previous trained gating ConvNet by jointly learning the gating fusion weights and the action classification. Different from \cite{convfusion16}, our work learns the gating fusion weights and action classification simultaneously in the gating ConvNet, while \cite{convfusion16} only has a classification branch. As shown in Table \ref{table3}, after adding a classification branch, the accuracy of the gated TSN increases by 0.08\%, 0.36\% and 0.7\% on the three splits of UCF101 respectively. So, it can be concluded that learning the gating fusion weights could benefit from learning the gating ConvNet for action classification.

\begin{table}[t]
\caption{Comparison of the accuracy (\%) among different fusion methods for different network architectures and different two-stream methods on the UCF101 (split 1).}
\begin{center}
\label{table4}
\vbox{
\renewcommand{\baselinestretch}{1.25}
{\footnotesize\centerline{\tabcolsep=10pt\begin{tabular*}{0.5\textwidth}{cccc}
\toprule
\tabincell{c}{Architectures } & gated fusion & weighted ave & SCI\\
\hline
CaffeNet (two-stream) &{\bf 71.80}&71.75&71.34\\
VGG16 (two-stream) &{\bf 78.94}&78.82&77.61\\
CaffeNet (TSN) &{\bf 74.93}&74.55&70.86\\
VGG16 (TSN) &{\bf 88.02}&87.95&85.00\\
BN-Inception (TSN) &{\bf 94.11}&93.81&93.96\\
\bottomrule
\end{tabular*}}}}
\end{center}
\end{table}

\begin{table}[t]
\caption{Accuracy (\%) of the gated TSN when the gating ConvNet does joint learning of action classification and learning the gating fusion weights  on the UCF101 (three splits).}
\begin{center}
\label{table3}
\vbox{
\renewcommand{\baselinestretch}{1.25}
{\footnotesize\centerline{\tabcolsep=10pt\begin{tabular*}{0.475\textwidth}{ccc}
\toprule
UCF101 split & gated fusion & \tabincell{c}{+gating ConvNet  classification} \\
\hline
\tabincell{c}{split 1}   &94.11&{\bf94.19}\\
\tabincell{c}{split 2}  &94.12&{\bf 94.48}\\
\tabincell{c}{split 3}  &94.14&{\bf 94.84}\\
\bottomrule
\end{tabular*}}}}
\end{center}
\end{table}

\subsection{Comparison with the State of the Art}

\begin{table}[t]
\caption{Comparison of the accuracy (\%) of our gated TSN with other state-of-the-art methods.}
\begin{center}
\vbox{
\renewcommand{\baselinestretch}{1.25}
{\footnotesize\centerline{\tabcolsep=10pt\begin{tabular*}{0.3\textwidth}{cc}
\toprule
Methods & UCF101\\
\hline
IDT \cite{IDT13} &               85.9\\
MoFAP \cite{MOFAP16}              &88.3\\
Two-stream ConvNet \cite{twostream14}  &88.0\\
C3D (3 nets) \cite{C3D15} &      85.2  \\
FstCN \cite{SunJYS15} &             88.1   \\
LTC \cite{LTC17} &               91.7 \\
ST-ResNet \cite{feichtenhofer2016spatiotemporal} &         93.4\\
TSN(2 modalities) \cite{TSN16}  & 94.0\\
TSN(3 modalities) \cite{TSN16}  & 94.2\\
{\bf Gated TSN(ours)} & {\bf 94.5}\\
DOVF \cite{DOVF17} & 94.9\\
TLE \cite{TLE16} & 95.6\\
\bottomrule
\end{tabular*}}}}
\label{Table6}
\end{center}
\end{table}

After above analysis of the gating ConvNet, final experiments on all three splits of UCF101 are implemented with our proposed methods. Specifically, the layers of inception4e from the two streams with conv fusion are taken as the inputs of the gating ConvNet. ReLU is chosen as the gating output activation function. Joint learning of the gating fusion weights and the gating ConvNet for action recognition is adopted. Mean average accuracy on three test sets of UCF101 is calculated as the final result. As shown in Table \ref{Table6}, the gated TSN are compared with both traditional approaches \cite{IDT13, MOFAP16} and deep learning methods \cite{twostream14, TSN16, LTC17, C3D15, TLE16, DOVF17, feichtenhofer2016spatiotemporal}. It is noted that our gated TSN only employs 2 modalities (RGB frame and optical flow stacks) as inputs and improves upon the original TSN with 2 modalities by 0.5\%. It even exceeds the TSN with 3 modalities by 0.3\%. This improvement demonstrates that the weighted averaging fusion with fixed weight could not fully exploit the capacity of the two streams on different samples, even with three streams, while the TSN with our gated fusion method could improve performance by adaptively assigning the fusion weights to different streams. It also exceeds traditional methods such as IDT \cite{IDT13}, MoFAP \cite{MOFAP16}, deep learning method such as LTC (Long-term temporal ConvNet)\cite{LTC17} ,C3D (Convolutional 3D)\cite{C3D15}, FstCN (Factorized ST-ConvNet)\cite{SunJYS15}, ST-ResNet (Spatio-Temporal ResNet)\cite{feichtenhofer2016spatiotemporal}. Though there are several methods that exceed the performance of the gated TSN, they also have some drawbacks. TLE (Temporal Linear Encoding) \cite{TLE16} consumes much more memory than ours due to the high dimension of bilinear pooling encoding \cite{bilinear2015,compactbilinear2015}. Lan {\sl et al}.\cite{DOVF17} stores off-line video features and then employs SVM for classifier. While being space-demanding, DOVF (Deep Local Video Feature) \cite{DOVF17} is less elegant than our gated TSN with end-to-end learning.

\subsection{Network Visualization}
After learning the gating fusion weights and the gating ConvNet for action classification jointly, visualizations of our networks are implemented on the validation set of UCF101 split 2. For each video, three equal spaced frames and corresponding optical flow stacks (with 5 frames) are sampled for the network inputs. The layers of inception4e from the two streams with conv fusion are taken as the inputs of the gating ConvNet.

\begin{figure*}[!t]
\centering
\includegraphics[width=1\linewidth]{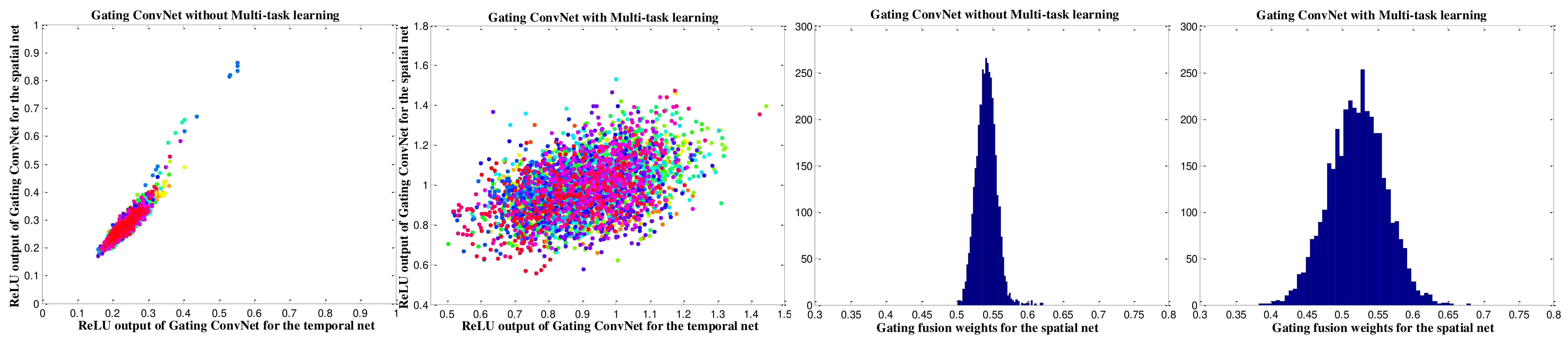}
\caption{The distribution of the fusion weights for the gated TSN (Top row), where horizontal axis and vertical axis represent the ReLU output of gating ConvNet for the temporal net and the spatial net respectively. Corresponding histograms of the fusion weights (Bottom row) for the spatial net where horizontal axis represents the gating fusion weights for the spatial net and vertical axis stands for the number of samples. All on the validation set of UCF101 (split 2).}
\label{Figure2}
\end{figure*}

In Fig. \ref{Figure2}, the distributions of the fusion weights for the gated TSN with and without our multi-task learning are displayed, corresponding histograms of the fusion weights for the spatial net are followed. None of the coordinate of the points on the first two subplots is zero, implying that the gating ConvNet has learnt that combining the spatial and the temporal nets is better than that only with single network or no network. With our multi-task learning, the output points of the gating ConvNet distribute more sparsely than that without multi-task learning. It could also be observed in the last subplot that the fusion weights for the spatial net range from 0.4 to 0.7. It is more wider than that without multi-task learning, whose fusion weights for the spatial net are mostly centered between 0.5 and 0.65. This may account for the 0.36\% increase of the accuracy on UCF101 split2 in Table \ref{table3}. With our joint learning method, an adaptive selection space is expanded for assigning the fusion weights for the two streams with more variations according to the current inputs.

\begin{figure*}[!t]
    \centering
    \includegraphics[width=1\linewidth]{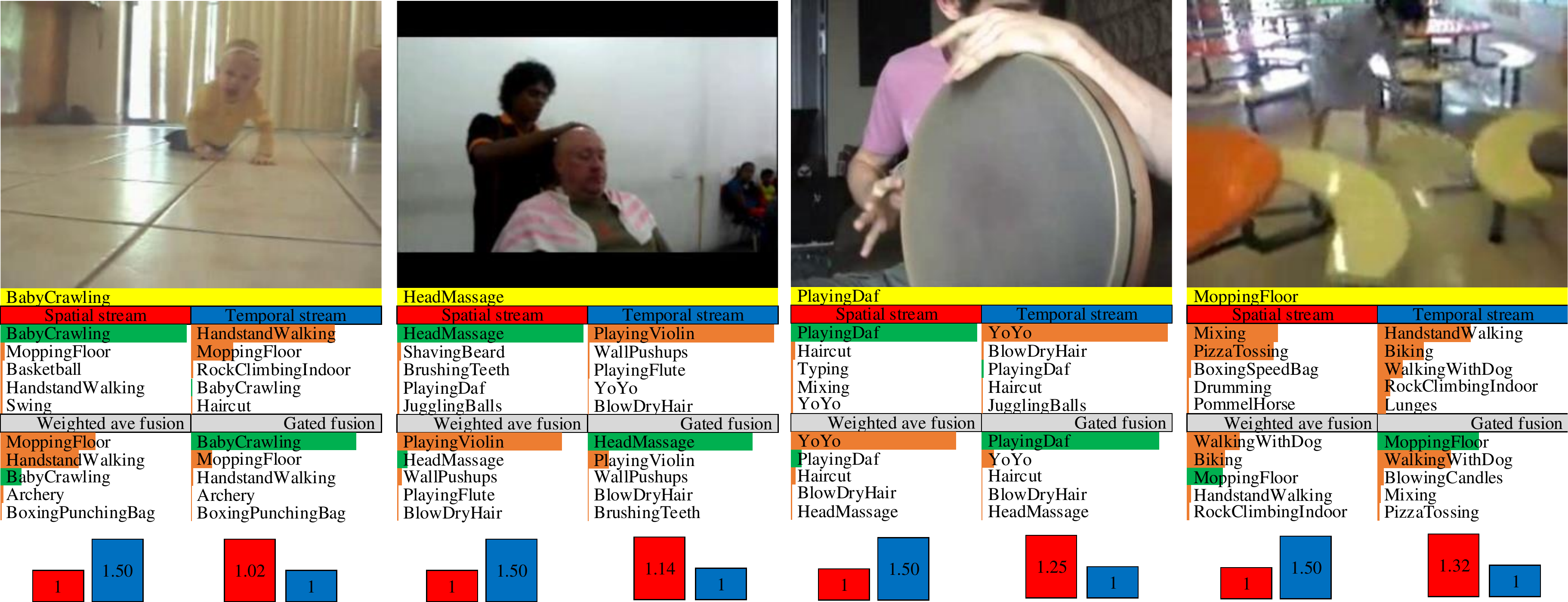}
    \caption{A comparison of top-5 predictions between the weighted averaging fusion with fixed weights and our gated fusion method on UCF101. The yellow bars stand for the ground truth label. The predictions of single stream are also shown, red for the spatial stream and blue for the temporal stream. Green and orange bars indicate correct and incorrect predictions respectively and the length of each bar shows its confidence. The predictions after different fusion methods and their fusion weights (confidence ratio) in the bottom of the figure by different fusion method are also displayed correspondingly. }
    \label{Figure4}
\end{figure*}

\begin{figure}[!t]
\centering
\includegraphics[width=1\linewidth]{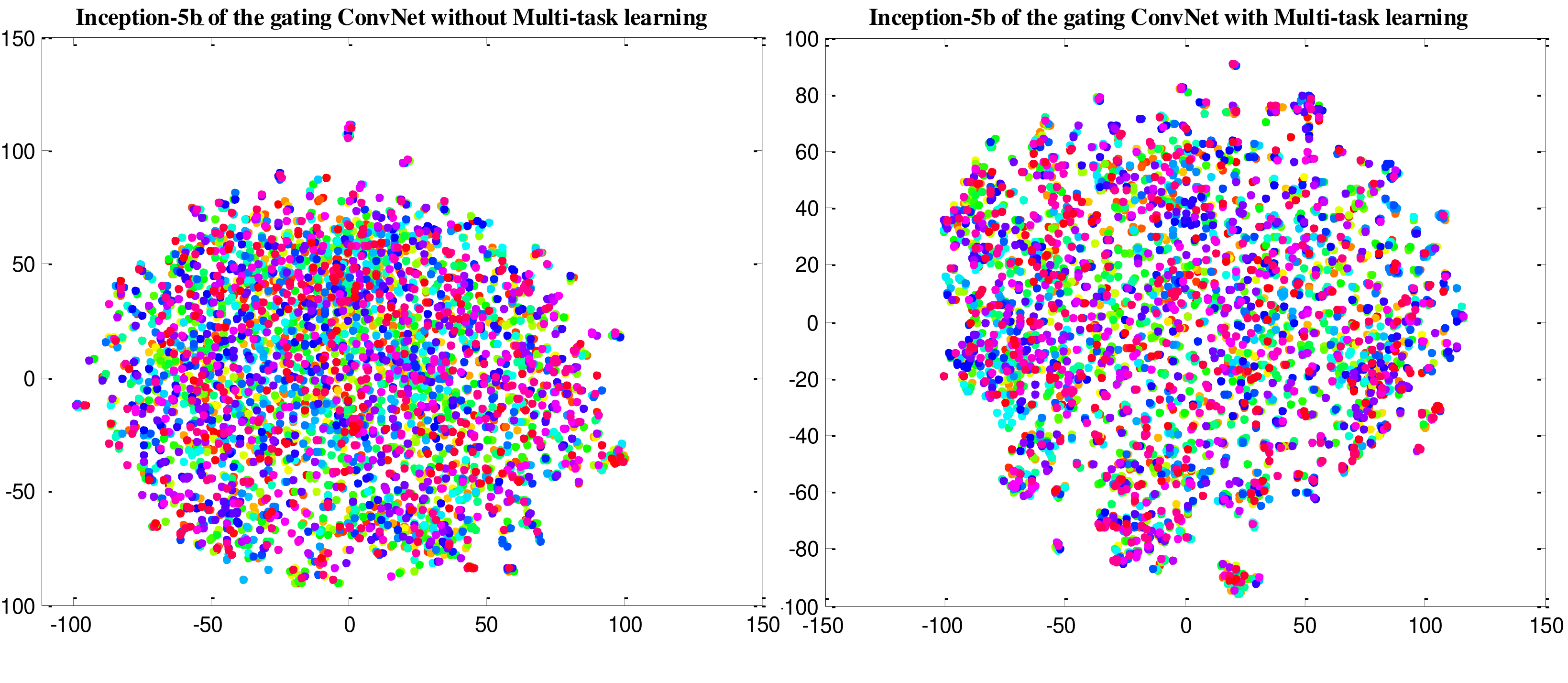}
\caption{t-SNE visualization of the features of the last convolutional layer of the gating ConvNet without and with our multi-task learning on the validation set of UCF101 (split 2). }
\label{Figure3}
\end{figure}

By t-SNE \cite{t-SNE08}, the features of the last convolutional layer of the gating ConvNet without and with our multi-task learning are visualized in Fig. \ref{Figure3}. The outputs of inception5b with 1024 dimension are mapped to the two dimensional points. After training with our multi-task learning approach, the features of the gating ConvNet become more discriminative than that without multi-task learning. By doing action classification, the features of the gating ConvNet could become more semantic, which distills knowledge into the learning of the fusion weights for the two streams. So, learning the gating fusion weights could benefit from learning action recognition in the gating ConvNet.

Finally, some examples of the classification results of the gated TSN and the weighted averaging fusion are shown and compared in Fig. \ref{Figure4}. As mentioned above, the fusion weights of TSN with the weighted averaging fusion is fixed for all samples. Though Wang {\sl et al}. \cite{TSN16} has already considered giving a higher weight to the temporal net (spatial vs temporal = 1 : 1.5) due to its higher accuracy than the spatial one, it could not fully exploit the capacity of the spatial and the temporal nets. In the first three subplots of Fig. \ref{Figure4}, the spatial stream always has high confidence for the ground truth label, while the temporal stream has high confidence for the incorrect class. In these cases, the higher fusion weights for the temporal stream than the spatial one may weaken the confidence to the ground truth, may even lead to prediction failures just as shown in these three examples. Our gated fusion assigns the spatial stream higher weights than the temporal one in all these three cases, which gives correct predictions with higher confidence, and proves it has learned that the spatial stream should be trusted more in these cases. It is also noticed that in the fourth subplot of Fig. \ref{Figure4}, the ground truth label, namely MoppingFloor, is not predicted into the top-5 by both the two streams, but after both fusion methods, it appears again. Fusing the predictions of the two streams with our gated fusion brings the result to be true by giving higher weights to the spatial stream than the temporal one.

\section{Conclusions}
In this work, we focus on learning the gating ConvNet for the two-stream ConvNets in action recognition. An end-to-end trainable gated fusion method is proposed and ReLU is adopted as the gating output activation function. Different input layers for the gating ConvNet and two different fusion methods (concatenation and conv fusion) are explored for these layers. Besides, it is shown that our joint learning of the gating fusion weights for the two streams and learning the gating ConvNet for action classification is helpful in improving the accuracy of the gated TSN. A high accuracy of 94.5\% is achieved on UCF101. In future work, we hope to learn the gating ConvNet to fuse the features of the two streams, which is different from the prediction level fusion in this paper. Our techniques in this work could also be extended to the semantic segmentation domain, where multi-stream deep neural networks are employed \cite{AdapNet17}.


\end{document}